%% file: main.tex
\documentclass[10pt,twocolumn,letterpaper]{article}

\usepackage{cvpr}              % To produce the CAMERA-READY version
\usepackage{soul}
\usepackage{cite}
\usepackage{amsmath,amssymb,amsfonts}
\usepackage{algorithmic}
\usepackage{graphicx}
\usepackage{textcomp}
\usepackage[dvipsnames]{xcolor}
\usepackage{booktabs}
\usepackage{graphicx}
\usepackage{subcaption} 

\usepackage{multirow}

\usepackage{colortbl}


\definecolor{cvprblue}{rgb}{0.21,0.49,0.74}
\usepackage[pagebackref,breaklinks,colorlinks,citecolor=cvprblue]{hyperref}
\usepackage[capitalize]{cleveref}
\crefname{section}{Sec.}{Secs.}
\Crefname{section}{Section}{Sections}
\Crefname{table}{Table}{Tables}
\crefname{table}{Tab.}{Tabs.}

\def\paperID{6} % *** Enter the Paper ID here
\def\confName{CVPR}
\def\confYear{2024}

\title{VT-Former: An Exploratory Study on Vehicle Trajectory Prediction for Highway Surveillance through Graph Isomorphism and Transformer}

\author{Armin Danesh Pazho\\
{University of North Carolina Charlotte}\\
{\tt\small adaneshp@uncc.edu}
\and
Ghazal Alinezhad Noghre\\
University of North Carolina Charlotte\\
{\tt\small galinezh@uncc.edu}
\and
Vinit Katariya\\
University of North Carolina Charlotte\\
{\tt\small vkatariy@uncc.edu}
\and
Hamed Tabkhi\\
University of North Carolina Charlotte\\
{\tt\small htabkhiv@uncc.edu}
}

\begin{document}
\maketitle
\input{tex/abstract}
\input{tex/introduction}
\input{tex/related_works}
\input{tex/methodology}

\input{tex/setup}
\input{tex/results}

\input{tex/ablation}
\input{tex/future}
\input{tex/conclusion}
% \section*{Acknowledgment}
% This research is supported by the National Science Foundation (NSF) under Award No. 1932524.
\section*{Acknowledgment}
This research is supported by the National Science Foundation (NSF) under Award No. 1932524.

{
    \small
    \bibliographystyle{ieeenat_fullname}
    \bibliography{main}
}


\end{document}

%% file: tex/abstract.tex
\begin{abstract}
Enhancing roadway safety has become an essential computer vision focus area for Intelligent Transportation Systems (ITS). As a part of ITS, Vehicle Trajectory Prediction (VTP) aims to forecast a vehicle's future positions based on its past and current movements. VTP is a pivotal element for road safety, aiding in applications such as traffic management, accident prevention, work-zone safety, and energy optimization. While most works in this field focus on autonomous driving, with the growing number of surveillance cameras, another sub-field emerges for surveillance VTP with its own set of challenges. In this paper, we introduce VT-Former, a novel transformer-based VTP approach for highway safety and surveillance. In addition to utilizing transformers to capture long-range temporal patterns, a new Graph Attentive Tokenization (GAT) module has been proposed to capture intricate social interactions among vehicles. This study seeks to explore both the advantages and the limitations inherent in combining transformer architecture with graphs for VTP. Our investigation, conducted across three benchmark datasets from diverse surveillance viewpoints, showcases the State-of-the-Art (SotA) or comparable performance of VT-Former in predicting vehicle trajectories. This study underscores the potential of VT-Former and its architecture, opening new avenues for future research and exploration.

\end{abstract}

%% file: tex/introduction.tex
\section{Introduction}
\label{sec:introduction}

Intelligent Transportation Systems (ITS) have emerged as a significant area of interest in research, with a notable classification distinguishing between autonomous driving \cite{cui2021uncertainty, xu2023improved, singh2023transformer, singh2023surround, fadadu2022multi, laddha2021mvfusenet, gautam2021sdvtracker, Nguyen_2022_CVPR, Djuric2018ShorttermMP, djuric2021multixnet, Liang_2022_CVPR} and surveillance applications \cite{nasirudin2023surveillance, kim2023cnn, putra2018proposed, yang2023cooperative, wang2023variant, dilek2023computer}. The subject of Vehicle Trajectory Prediction (VTP), as a subset of ITS, has garnered significant attention in both areas. VTP is utilized within various higher-level applications among which are traffic management and control \cite{katariya2022deeptrack, rossi2021vehicle, liang2021nettraj, lu2019trajectory, wang2020improving}, accident and collision avoidance \cite{houenou2013vehicle, kang2017parametric, lyu2020vehicle}, workzone safety \cite{katariya2022deeptrack, weng2014analysis, wu2019modified}, route planning \cite{rathore2019scalable, xia2017uptp}, and vehicle anomaly detection \cite{peralta2023outlier, karatacs2021trajectory, huang2021data, jiao2023learning}, etc.

\begin{figure}[]
    \centering
    % \resizebox{1\linewidth}{!}{
    \includegraphics[clip,trim={17 18 23 18},width=1\columnwidth]{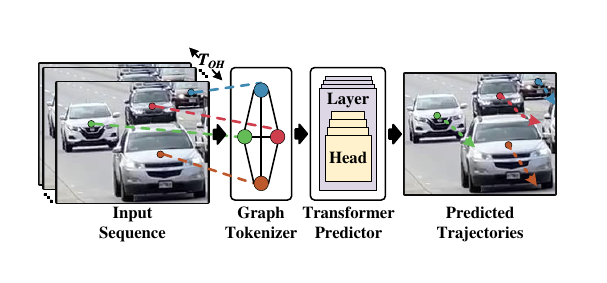}
    % }
    \caption{A conceptual overview of the VT-Former. The input sequence, characterized by an observation horizon denoted as $T_{OH}$, undergoes tokenization via a novel graph-based tokenizer. The resultant tokens are subsequently processed through a transformer-based predictive model to forecast future trajectories.}
    \label{fig:intro}
\end{figure}

The rapid advancements in computer vision and the widespread availability of surveillance cameras have opened up promising avenues and introduced novel techniques for advancing Surveillance Vehicle Trajectory Prediction (SVTP) methods \cite{cui2019multimodal, huang2022survey, leon2021review}. This study focuses on SVTP in which the historical vehicle movement is used to forecast the future path, position, and behavior of a vehicle over a specific time horizon. While many studies focus on trajectory prediction from the perspective of autonomous driving and autonomous vehicles \cite{tang2023trajectory, hsu2023deep, bhatt2023mpc, mcallister2022control}, SVTP models approach it from surveillance angles, including the bird's eye view, high-angle, and eye-level perspectives \cite{alinezhad2023pishgu, katariya2022deeptrack}. Another notable distinction between the two domains is that models in autonomous driving often predict multiple trajectories and select the most suitable one for evaluation \cite{messaoud2020attention, cui2019multimodal}. In contrast, SVTP methods typically rely on deterministic models that forecast a singular trajectory \cite{alinezhad2023pishgu, katariya2022deeptrack, li2019grip++}, which is more aligned with the practical requirements of highway safety scenarios.

There are several challenges and requirements associated with SVTP. Many applications such as traffic management or vehicle anomaly detection, require a balance between efficiency and accuracy \cite{messaoud2020attention, li2023amgb, alinezhad2023pishgu}. This balance is pivotal as it directly impacts the real-world applicability and effectiveness of SVTP models in dynamic and often unpredictable environments. Another challenge is that vehicles interact with each other. Predicting how these interactions will unfold is complex, especially when combined with the uncertain dynamics of vehicle movements. Furthermore, a significant challenge stems from the design of most SVTP approaches for a bird's-eye view, whereas highway surveillance typically employs high-angle view cameras instead of bird's-eye views. Therefore, it is crucial to assess such models from the perspective of these more common real-world viewpoints as well.

In this paper, we present VT-Former, an innovative approach that leverages the power of transformers for vehicle trajectory prediction for highway surveillance applications. VT-Former also introduces Graph Attentive Tokenization (GAT) to capture the underlying social interactions among moving vehicles within the scene. The use of graphs has demonstrated significant success in pedestrian trajectory prediction, as evidenced by prior studies \cite{mendieta2021carpe, alinezhad2023pishgu}. This exploratory study aims to assess their effectiveness in the vehicle surveillance domain. Integrating these key components, GAT, and the decoder-only transformer, within VT-Former, offers a novel approach for SVTP. \cref{fig:intro} shows a conceptual overview of VT-Former.

To demonstrate the benefits and limitations of VT-Former, we perform extensive experimentation using three major datasets: NGSIM \cite{NGSIM_i80, NGSIM_US101}, CHD High-angle and CHD Eye-level \cite{chd} datasets, to comprehensively evaluate the performance of the VT-Former model from diverse perspectives. The evaluation encompassed three distinct angle variations corresponding to each dataset. Please note that VT-Former is designed for highway surveillance, not autonomous driving, making datasets like Argoverse \cite{2021_4734ba6f, Chang_2019_CVPR}, nuScenes \cite{Caesar_2020_CVPR}, PUP \cite{ivanovic2022heterogeneous}, and Waymo \cite{Sun_2020_CVPR, Ettinger_2021_ICCV} unrelated to this work. The outcomes underscore the advantages and potential, as well as the shortcomings of employing GAT alongside transformers for enhancing trajectory prediction accuracy and efficiency. On top of comparison with SotA models, we delved into a thorough analysis, comparing the effects of shortened historical trajectory data against extended historical horizons. This analysis explores the benefits and shortcomings of using a shortened observation horizon in cases where quick response time or lower prediction latency is critical. Moreover, our ablation study demonstrates the advantages of utilizing a higher frame rate over the conventional setup, while preserving the existing ratio between the prediction horizon and the observed trajectory.

Overall, the contributions of this paper can be summarized as follows:

\begin{itemize}
    \item Introducing VT-Former, exploring the combination of graph technology and transformer architecture as a new avenue and a novel approach for vehicle trajectory prediction for highway safety applications.
    \item Comprehensive assessments on benchmark datasets encompassing diverse surveillance camera angles, including bird's-eye view, high-angle view, and eye-level view.
    \item Exploring the influence of reduced observation horizons as well as higher frame rate on model accuracy, underscoring its effectiveness in scenarios demanding rapid trajectory prediction.
    % \item Assessing the real-time efficiency of the model by executing it on an embedded platform.
    % \item Demonstrating the practical utility of VT-Former by implementing a straightforward driving anomaly detection algorithm and evaluating it on adversarially generated anomaly scenarios.
\end{itemize}

%% file: tex/related_works.tex
\section{Related Works}
\label{sec:related}

In recent years, ITS have emerged as a focal point of extensive research endeavors within the scientific community \cite{al2023methods, 10268597, fainekos2023systems, michael2023artificial, tiausas2023hprop}. Modern systems are increasingly integrating deep learning-based prediction mechanisms \cite{olowononi2020resilient, alinezhad2023pishgu, liu2019edge}. For many applications, path prediction is crucial as it helps in foreseeing future actions and locations beneficial for many downstream tasks. These predictive tools normally aim to operate in real-time \cite{Isern2020Reconfig, katariya2022deeptrack}. Such methodologies can be utilized for forecasting traffic densities \cite{Jeong2013traffFlow, chen2017research}, improving the safety of work zones \cite{sabeti2021toward}, and promoting autonomous vehicle technologies \cite{salzmann2020trajectron}. These predictive models also play a critical role in enabling advancements in smart monitoring and surveillance for enhanced security \cite{ramasamy2022secure}, improving pedestrian and worker safety in urban environments \cite{mendieta2021carpe, wang2022intelligent}, facilitating the identification and forecasting of anomalies within complex systems \cite{markovitz2020graph}, and optimizing the design of transportation routes for increased efficiency and resource allocation \cite{bradley2015optimization}.

\begin{figure*}[t]
    \centering
    % \resizebox{1\linewidth}{!}{
    \includegraphics[width=0.7\linewidth,clip,trim={18 18 35 17}]{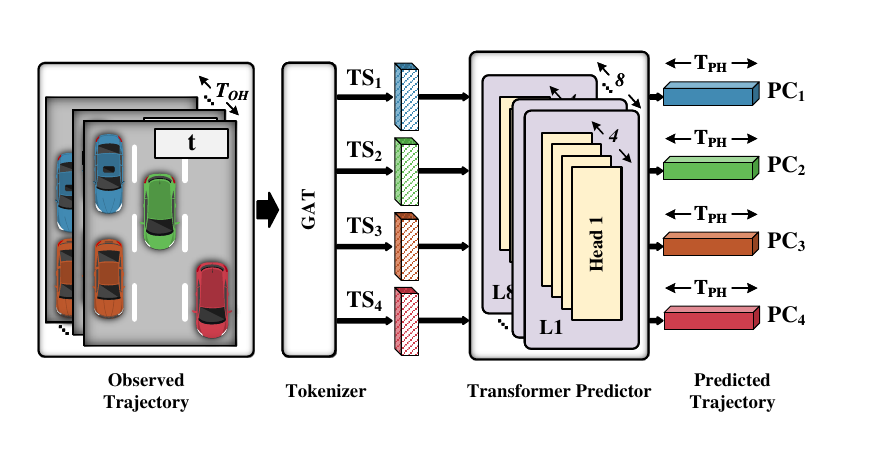}
    % }
    \caption{VT-Former at a glance: our approach begins with the application of Graph Attentive Tokenization (GAT) to create enriched token sequences that encapsulate social interactions. Subsequently, these enhanced sequences are fed through the Transformer Prediction module, which consists of 8 layers of transformer decoders, each equipped with 4 heads, generating predicted trajectories. VT-Former is an autoregressive sequence-to-sequence model that generates the output in multiple steps. However, for easier visualization, the output tokens are shown in the last step where they are completely generated. $T_{OH}$, $TS_i$, $T_{PH}$, and $PC_i$ denote the observation horizon time, token sequence of the $i^{th}$ vehicle, prediction horizon time, and the predicted trajectory, respectively.}
    \label{fig:overview}
\end{figure*}

Traditional methods, such as those based on Long Short-Term Memory (LSTM) networks, have been explored by Alahi et al. \cite{alahi2016social}. While effective, these methods often lack the flexibility and scalability of newer architectures. Graph Neural Networks (GNNs) have emerged as a powerful tool for trajectory prediction \cite{li2019grip, deo2022multimodal}, capitalizing on their ability to model relational data and complex interactions between entities. Specifically, Graph Convolutional Networks (GCN) \cite{cheng2023gatraj, salzmann2020trajectron} allows for localized feature learning on graphs, making them apt for capturing spatial relationships between vehicles \cite{lv2023ssagcn, liu2022multi}. The Graph Isomorphism Network (GIN) \cite{mendieta2021carpe} further extends this capability by distinguishing different graph structures, enabling more precise trajectory predictions by considering nuanced vehicular interactions \cite{xu2018powerful,velivckovic2017graph,kipf2016semi}.

In recent studies, there has been a notable shift from Long Short-Term Memory (LSTM) and Convolutional Neural Networks (CNNs) to Transformers, as introduced by Vaswani et al. \cite{vaswani2017attention}. Transformers employ self-attention mechanisms for the analysis of sequential data making them suitable for many applications such as computer vision \cite{10.1145/3505244} and natural language processing \cite{sutskever2014sequence}. Simultaneously, there have been endeavors to harness the power of Transformers in predicting trajectories within the realm of autonomous vehicles, enabling a better understanding of the temporal evolution of object movements \cite{zhang2022spatTemp, jia2023HDGT}. However, transformer-based methods remain largely unexplored within highway surveillance and safety applications. Moreover, their potential synergy with graphs to more effectively capture social nuances has yet to be investigated.

Despite the multitude of trajectory prediction algorithms proposed over the past decade, a limited number are explicitly designed for highway environments \cite{altche2017lstm, katariya2022deeptrack}. Furthermore, the field of highway trajectory prediction is hampered by the scarcity of comprehensive datasets. Some recent works such as \cite{chd} have attempted to mitigate this and fill the gap. However, there remains a distinct requirement for specialized datasets tailored to highway-based applications. In general, there is a noticeable disparity between research endeavors and real-world applications in trajectory prediction especially when using already installed surveillance cameras. This underscores the potential for research and further development in this domain.

%% file: tex/methodology.tex
\section{VT-Former Formulation and Methodology}

At its core, VT-Former is a transformer-based approach for SVTP in roadways. Transformers are renowned for their exceptional generalization capabilities  \cite{vaswani2017attention}, enabling them to discern intricate long-term patterns within data and excel in tasks featuring extensive and diverse datasets. While the original Transformer model was computationally demanding and exhibited slower performance as it had both encoder and decoder components, numerous optimizations and model adaptations have been developed to enhance their efficiency \cite{kenton2019bert, brown2020language} by choosing only encoder or decoder as the core of their models. This has facilitated the utilization of transformers in applications necessitating low-latency responses. These attributes motivated our choice of the decoder-only transformer \cite{brown2020language} as the core component of our algorithm.

VT-Former model utilizes a temporal window of the movement of a vehicle as input. This window is precisely determined by the spatial coordinates denoting the center of the vehicle's bounding box and extends across the entirety of the observation horizon, creating a temporal window of the movements of the vehicle along the time axis:

\begin{equation}
    C_i = [(x_i, y_i)^{t-T_{OH}+1}, (x_i, y_i)^{t-T_{OH}+2}, ..., (x_i, y_i)^t]
\end{equation}

where $(x_i, y_i)^t$ represents the center of the $i^{th}$ vehicle's bounding box and $T_{OH}$ is the observation horizon. The model predicts the future location of vehicles through the prediction horizon ($T_{PH}$):

\begin{equation}
    PC_i = [(x'_i, y'_i)^{t+1}, (x'_i, y'_i)^{t+2}, ..., (x'_i, y'_i)^{t+T_{PH}}]
\end{equation}

The overview of VT-Former can be seen in \cref{fig:overview}. Our proposed design consists of two main modules; the Graph Attentive Tokenization (GAT) module and the Transformer Prediction module. In the following subsections, we will delve into the details of the proposed architecture.

\begin{figure*}[ht!]
    \centering
    % \resizebox{1\linewidth}{!}{
    \includegraphics[clip,trim={17 22 18 18},width=1\textwidth]{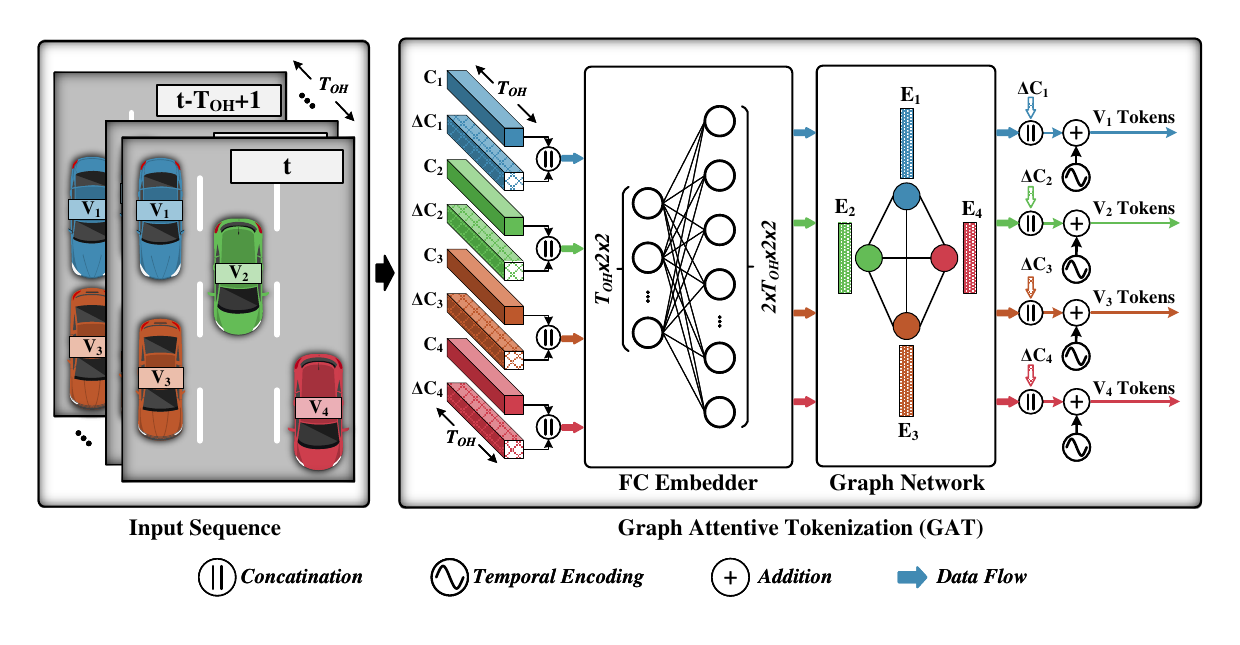}
    % }
    \caption{Graph Attentive Tokenization (GAT). GAT gets the past trajectory ($C_i$) of each vehicle ($V_i$) over the observation horizon ($T_{OH}$) for each vehicle and calculates the relative movement ($\Delta C_i$). These two components are concatenated and passed through a fully connected layer, expanding the feature maps by a factor of 2. Subsequently, we employ a Graph Network to capture intricate interactions among vehicles within the scene. The output of this Graph Network is further concatenated with the relative movement, accentuating the temporal evolution within the sequence. Finally, to infuse temporal order into the token sequences, we leverage temporal encoding.}
    \label{fig:token}
\end{figure*}

\subsection{Graph Attention Tokenization}
\label{sec:token}
Graph Attention Tokenization (GAT) illustrated in \cref{fig:token} aims to extract enriched tokens that encapsulate the social interactions between the vehicles available in the scene. In order to accentuate the relative movements of the vehicles through the observation window we define the relative trajectory with respect to the first observed location:
\begin{equation}
\begin{split}
       \Delta C_i = [(x_i, y_i)^{t-T_{OH}+1}-(x_i, y_i)^{t-T_{OH}+1}, \\
       (x_i, y_i)^{t-T_{OH}+2}-(x_i, y_i)^{t-T_{OH}+1},\\
       ..., (x_i, y_i)^t-(x_i, y_i)^{t-T_{OH}+1}]
\end{split}
\end{equation}

$C_i$ and $\Delta C_i$ are concatenated to form the input of the fully connected layer which expands the dimension of the input features by a factor of $2$. In the next step, we leverage Graph Isomorphism Network (GIN) \cite{xu2018powerful} to capture the relationships between subjects available in the scene. GINs have proven highly effective in the context of processing graph-structured data, where nodes and edges signify entities and their relationships, respectively. In our formulation, each vehicle will be considered as a node in a fully connected graph. We chose to have a fully connected graph so that the Graph Network does not have any predefined biases and all relationships are learned by training. Another benefit of having a fully connected graph is that by performing just one step graph operation, the information will be propagated over all nodes and further steps are not necessary leading to a more efficient design. Inspired by \cite{alinezhad2023pishgu}, we design separate aggregation networks for the node feature and the neighbor features. The node feature aggregator is an MLP network with one hidden layer:

\begin{equation}
    E'_v = Act(W_1 \cdot Act(W_0 \cdot(\beta\cdot E_v) +B_0) + B_1)
\end{equation}

where $E_v$ is the feature map for vehicle $v$, $\beta$ is a learnable parameter, and $W_0$, $B_0$, $W_1$ and $B_1$ are the parameters of an MLP with one hidden layer. We have used the Leaky ReLu function as an activation function. 

For aggregating neighbor features, we designed a more sophisticated network equipped with an attention mechanism: 

\begin{equation}
\begin{split}
E'_u = Act(W_2 \cdot E_u + B_2 )\\
E''_u = A_s[A_C (E'_u) \times E'_u] \times [A_C (E'_u) \times E'_u ] + E'_u \\
\text{where u}  \in \mathcal{N}(v)
\end{split}
\end{equation}

$E_u$ is the feature map of neighbour vehicle, $\mathcal{N}(v)$ is the set of neighbour vehicles, $W_2$ and $B_2$ are the parameters of a fully connected layer, $Act$ is Leaky ReLu function, and $A_s$ and $A_c$ are spatial attention and channel attention mechanism introduced by \cite{woo2018cbam}. The attention mechanism helps the network to focus more on the features from neighbors that are most informative for predicting the future trajectory. The sum of $E'_v$ and $E''_u$ for all neighbor nodes is again concatenated with $\Delta C_i$ to emphasize relative movement constructing the predictor input sequence. In order to also add temporal ordering to the tokens in the predictor input sequence, we leverage temporal encoding proposed by \cite{vaswani2017attention} as depicted by \cref{fig:token}. This process is necessary due to the fact that the Transformer Predictor module in the next step is not able to grasp the order of the sequence as suggested by \cite{vaswani2017attention}.

\subsection{Transformer Predictor Module}
In the domain of time series data processing, Recurrent Neural Networks (RNNs) have traditionally been the primary choice. However, RNNs exhibit limitations related to their short-term memory, often constrained by challenges like vanishing gradients, which limits their ability to capture relationships between distant elements within a sequence. Several works also have tried using variations of Convolutional Neural Networks (CNNs) \cite{alinezhad2023pishgu, mendieta2021carpe} for finding temporal patterns, however, CNNs are not the preferred choice for modeling time series data because they lack the ability to effectively capture temporal dependencies, handle variable sequence lengths, and recognize long-range sequential patterns due to limited kernel size. Recently, a substantial paradigm shift has occurred with the emergence of transformers \cite{vaswani2017attention}. Transformers have garnered significant attention, predominantly due to their capability to capture long-range dependencies. Originally designed for Natural Language Processing (NLP) tasks, transformers have been adopted across a vast number of domains, including but not limited to computer vision, image and video processing, etc.

Hence, we have adopted a transformer-based architecture for more accurate vehicle trajectory prediction. The attention score inside the transformer layers is calculated using three learnable matrices Query (Q), Key (K), and Value (V) as described in \cite{vaswani2017attention}:

\begin{equation}
\label{attention}
\operatorname{Attention}(Q, K, V)=\operatorname{softmax}\left(\frac{Q K^T}{\sqrt{d_k}}\right) V
\end{equation}

where $d_k$ is the dimensionality of the Key matrix. Multiple attention heads are employed in transformers making them capable of attending to various input patterns. In our network, we have opted for a configuration with four attention heads. Following the attention mechanism, a Feed Forward Network (FFN) with a single hidden layer is utilized:

\begin{equation}
\operatorname{FFN}(x)=\operatorname{Act}\left(x \cdot W_3+b_3\right) \cdot W_4+b_4
\end{equation}

where $W_3$, $b_3$, $W_4$, and $b_4$ are the parameters of the two fully connected layers.

The Transformer Predictor Module consists of 8 decoder layers stacked on top of each other. We chose the dimension of the FFN to be a size of 256. Within our design, as detailed in \cref{sec:token}, we incorporated inter-vehicle attention in the tokenization process, although temporal attention among tokens was not considered. The multi-head attention mechanism in transformers is an ensemble of distinct attention mechanisms, with each head specializing in particular patterns. This adaptability empowers transformers to effectively capture a wide spectrum of data dependencies in the temporal dimension, ultimately improving the accuracy of predicted trajectories. We choose an autoregressive strategy where the model generates the sequence of predicted locations one step at a time in an ordered manner, where each prediction is influenced by the previous predictions leading to more accurate prediction. To prevent the Transformer Predictor from peeking into future information, we employ future masking while training. The masking procedure is particularly important to ensure the causality of the system in parallel training mode.

%% file: tex/setup.tex
\section{Experimental Setup}

All trainings have been conducted on a server equipped with 2x EPYC 7513 processors and 3x Nvidia A6000 GPUs with 256GB of memory.

\subsection{Datasets}
\label{sec:datasets}

This study employs three datasets, each reflecting a distinct surveillance viewpoint (bird's-eye, high-angle, and eye-level), to evaluate the performance and robustness of the VT-Former model.

\subsubsection{NGSIM}
The NGSIM dataset, specifically the US-101 \cite{NGSIM_US101} and I-80 \cite{NGSIM_i80} subsets, offer a rich source of real-world traffic scenarios and driver behaviors. As suggested by the name, this dataset is gathered from two distinct locations namely southbound US 101 in Los Angeles, CA, and eastbound I-80 in Emeryville, CA. With millions of data samples captured in a bird's-eye view at 10 Hz, it provides a comprehensive perspective of the surrounding environment. NGSIM data is originally collected from cameras positioned on buildings near the freeways, and it is then converted into a bird's-eye format. The bounding box and tracking information are all hand-annotated and thus the noise is minimal in this dataset. This holds particular significance in practical, real-world scenarios, especially when dealing with pre-installed surveillance camera systems where manually annotated data is absent. Consequently, transitioning from training on hand-annotated data to operational deployment in the real world may result in a degradation in the model's performance.

\subsubsection{Carolina's Highway Dataset}

Carolina's Highway Dataset (CHD) \cite{chd} is a comprehensive collection of vehicle trajectory data extracted from videos recorded in various locations across North Carolina and South Carolina, USA. It includes two distinct viewpoints: eye-level (CHD Eye-level) and high-angle (CHD High-angle), each included in a separate dataset. CHD not only provides trajectory information but also includes annotations for multi-vehicle detection and tracking. The data was recorded in full HD 60Hz at eight different locations, each chosen to capture a wide range of traffic patterns, vehicle behaviors, and lighting conditions. CHD offers a diverse perspective on incoming traffic, making it a valuable resource for understanding real-world highway scenarios. The annotations for CHD Eye-level and CHD High-angle datasets are derived through the application of deep learning models. Owing to the machine-based annotation process, these datasets exhibit a greater amount of noisy data in comparison to the NGSIM dataset. However, in contrast to NGSIM, these machine-generated annotations readily align with conditions and data obtainable in real-world settings.

\subsection{Metrics}
In alignment with previous studies, three established metrics, namely Average Displacement Error (ADE), Final Displacement Error (FDE), and Root Mean Squared Error (RMSE), are utilized to assess VT-Former's performance.

\subsubsection{Average Displacement Error (ADE)} ADE is a metric commonly used in the field of trajectory forecasting and motion prediction. It quantifies the accuracy of predicted trajectories for moving objects, such as vehicles or pedestrians by measuring the average Euclidean distance between the ground truth coordinates ($x^t, y^t$) and the predicted coordinates ($x'^t, y'^t$) across a defined prediction horizon ($T_{PH}$) considering all available $N$ vehicles in the scene:

\begin{equation}
ADE=\frac{\sum_{i=1}^N \sum_{t=1}^{T_{PH}} \sqrt{\left(x^t-x'^t\right)^2+\left(y^t-y'^t\right)^2}}{N \times T_{PH}} 
\end{equation}

In essence, it gauges how closely the predicted trajectory aligns with the actual trajectory, providing a numerical assessment of the model's forecasting performance. A lower ADE value indicates more accurate predictions, while a higher value suggests less accurate or less precise trajectory forecasts. ADE is a fundamental measure in evaluating the quality and reliability of trajectory prediction models.

\subsubsection{Final Displacement Error (FDE)} FDE quantifies the Euclidean distance between the ground truth final coordinates ($x^{T_{PH}}, y^{T_{PH}}$) and the predicted final coordinates ($x'^{T_{PH}}, y'^{T_{PH}}$) averaged over all available $N$ vehicles: 
\begin{equation}
FDE=\frac{\sum_{i=1}^N \sqrt{\left(x^{T_{PH}}-x'^{T_{PH}}\right)^2+\left(y^{T_{PH}}-y'^{T_{PH}}\right)^2}}{N} 
\end{equation}

In essence, FDE assesses how accurately the model can forecast the precise destination or concluding point of an object's trajectory. A smaller FDE value indicates more accurate predictions, signifying that the model successfully anticipates the final position, while a larger FDE value indicates less precise forecasting, implying a greater deviation from the actual endpoint.

\subsubsection{Root Mean Squared Error (RMSE)} RMSE measures the square root of the average of the squared differences between the ground truth trajectory and the predicted trajectory at time step $t$ over all available $N$ vehicles:

\begin{equation}
RMSE^t = \sqrt{\frac{1}{N} \sum_{i=1}^N\left[\left(x^t-x'^t\right)^2+\left(y^t-y'^t\right)^2\right]}
\end{equation}

A smaller RMSE indicates that the model's predictions are closer to the actual values, while a larger RMSE signifies a greater level of error in the predictions.

\subsection{Training Strategy and Hyperparameters}
\label{sec:strat}

In line with the conventions of previous research, our study utilizes an observation horizon of 15 timesteps and a prediction horizon of 25 timesteps (referred to as VT-Former$_{LH}$, where LH denotes Long History) across all datasets examined. Additionally, we define two more variations of VT-Former: the first version observes 10 timesteps and predicts 25 (VT-Former$_{MH}$, where MH stands for Medium History), and the second version observes 5 timesteps and predicts 25 (VT-Former$_{SH}$, where SH stands for Short History).

Given the disparate recording rates of these datasets, a standardization process is undertaken to achieve comparability with previous research, necessitating the down-sampling of both datasets to a common frame rate of 5 frames per second. For training the model for all datasets, Mean Squared Error (MSE) loss has been utilized.

Throughout all training sessions, the models undergo 80 training epochs, with a learning rate of 0.01, a weight decay of 0.0005, and a dropout rate of 0.2 consistently applied using Adam optimizer with batch size set to 16.

%% file: tex/results.tex
\section{Results} \label{sec:result}

\begin{table}[]
\centering
\caption{Comprasion with SotA methods (CS-LSTM \cite{deo2018convolutional}, GRIP++ \cite{li2019grip++}, STA-LSTM \cite{lin2021vehicle}, DeepTrack \cite{katariya2022deeptrack}, and Pishgu \cite{alinezhad2023pishgu}) on NGSIM \cite{NGSIM_US101, NGSIM_i80}. VT-Former results are shown with different observation horizons of 15, 10, and 5 for VT-Former$_{LH}$, VT-Former$_{MH}$, and VT-Former$_{SH}$ respectively. The results are in meters.}
\label{tab:ngsim3}
\resizebox{\columnwidth}{!}{%
\begin{tabular}{@{}ccc|ccccc|c@{}}
\toprule
                                        &              &              & \multicolumn{5}{c|}{\textbf{RMSE}}                                  & \multirow{2}{*}{\textbf{\begin{tabular}[c]{@{}c@{}}Params\\ (k)\end{tabular}}} \\ \cmidrule(r){1-8}
\multicolumn{1}{c|}{\textbf{Model}}     & \textbf{ADE} & \textbf{FDE} & \textbf{1s} & \textbf{2s} & \textbf{3s} & \textbf{4s} & \textbf{5s} &                                                                                \\ \midrule
\multicolumn{1}{c|}{\textbf{CS-LSTM}}   & 2.29         & 3.34         & 0.61        & 1.27        & 2.09        & 3.10        & 4.37        & 191                                                                               \\
\multicolumn{1}{c|}{\textbf{GRIP++}}    & 2.01         & 3.25         & 0.38        & 0.89        & 1.45        & 2.14        & 2.94        & -                                                                               \\
\multicolumn{1}{c|}{\textbf{STA-LSTM}}  & 1.89         & 3.16         & 0.37        & 0.98        & 1.17        & 2.63        & 3.78        & 124                                                                               \\
\multicolumn{1}{c|}{\textbf{DeepTrack}} & 2.01         & 3.21         & 0.47        & 1.08        & 1.83        & 2.75        & 3.89        & 109                                                                               \\
% \multicolumn{1}{c|}{\textbf{Pishgu}}    & 0.88         & 1.96         & 0.15        & 0.46        & 0.82        & 1.25        & 1.74        & 132                                                                               \\ \midrule
\midrule
\rowcolor[HTML]{D9D9D9}
\multicolumn{1}{c|}{\textbf{VT-Former$_{LH}$}} & 2.53         & 5.51         & 0.46        & 1.33        & 2.35        & 3.50        & 4.80        &  155   \\ 
\rowcolor[HTML]{D9D9D9}
\multicolumn{1}{c|}{\textbf{VT-Former$_{MH}$}} & 2.24         & 5.15         & 0.37        & 1.13        & 2.10        & 3.26        & 4.62        &  141    
\\
\rowcolor[HTML]{D9D9D9}
\multicolumn{1}{c|}{\textbf{VT-Former$_{SH}$}} & 2.10         & 4.91         & 0.30        & 0.99        & 1.90        & 3.00        & 4.31        &  132                                                                             \\ \bottomrule
\end{tabular}%
}
\end{table}

\begin{table}[]
\centering
\caption{Comparison with SotA methods on CHD \cite{chd} High-angle dataset. VT-Former results are shown with different observation horizons of 15, 10, and 5 for VT-Former$_{LH}$, VT-Former$_{MH}$, and VT-Former$_{SH}$ respectively. The results are in pixels.}
\label{tab:chd_h3}
\resizebox{\columnwidth}{!}{%
\begin{tabular}{@{}ccc|ccccc@{}}
\toprule
\multicolumn{1}{l}{}   & \multicolumn{1}{l}{} & \multicolumn{1}{l}{} & \multicolumn{5}{|c}{\textbf{RMSE}}                                   \\ \midrule
\textbf{Model}         & \textbf{ADE}         & \textbf{FDE}         & \textbf{1s} & \textbf{2s} & \textbf{3s} & \textbf{4s} & \textbf{5s} \\
\midrule
\textbf{S-STGCNN \cite{mohamed2020social}} & 31.87                & 98.46                & 9.74        & 21.83       & 29.01       & 42.34       & 82.14       \\
\textbf{GRIP++ \cite{li2019grip++}}        & 36.32                & 100.89               & 3.40         & 6.67        & 14.32       & 28.02       & 123.04      \\
\textbf{Pisghu \cite{alinezhad2023pishgu}}        & 18.33                & 61.92                & 4.04        & 7.48        & 13.99       & 24.30        & 51.51       \\
\midrule \rowcolor[HTML]{D9D9D9} 
\textbf{VT-Former$_{LH}$}          & 25.95                & 87.21                & 7.60         & 17.35       & 22.90        & 27.97       & 66.39       \\ 
\rowcolor[HTML]{D9D9D9} 
\textbf{VT-Former$_{MH}$}          & 25.90                      & 87.90                      & 6.40                    & 14.62                      & 20.56                    & 29.44                     & 70.05       \\ 
\rowcolor[HTML]{D9D9D9} 
\textbf{VT-Former$_{SH}$}          & 25.33                     & 88.99                     & 5.67                    & 12.96                     & 19.12                    & 29.83                     & 70.72       \\

\bottomrule
\end{tabular}%
}
\end{table}

\begin{table}[]
\centering
\caption{Comparison with SotA methods on CHD \cite{chd} Eye-level dataset. VT-Former results are shown with different observation horizons of 15, 10, and 5 for VT-Former$_{LH}$, VT-Former$_{MH}$, and VT-Former$_{SH}$ respectively. The results are in pixels.}
\label{tab:chd_e3}
\resizebox{\columnwidth}{!}{%
\begin{tabular}{@{}ccc|ccccc@{}}
\toprule
\multicolumn{1}{l}{\textbf{}} & \multicolumn{1}{l}{\textbf{}} & \multicolumn{1}{l}{\textbf{}} & \multicolumn{5}{|c}{\textbf{RMSE}}                                   \\ \midrule
\textbf{Model}                & \textbf{ADE}                  & \textbf{FDE}                  & \textbf{1s} & \textbf{2s} & \textbf{3s} & \textbf{4s} & \textbf{5s} \\\midrule
\textbf{S-STGCNN \cite{mohamed2020social}}        & 24.33                         & 95.22                         & 4.32        & 9.15        & 15.93       & 29.05       & 68.32       \\
\textbf{GRIP++ \cite{li2019grip++}}               & 44.27                         & 129.58                        & 4.42        & 12.86       & 24.31       & 35.04       & 145.17      \\
\textbf{Pisghu \cite{alinezhad2023pishgu}}               & 37.99                         & 123.69                        & 4.98        & 13.58       & 26.61       & 50.31       & 106.45      \\
\midrule 
\rowcolor[HTML]{D9D9D9} 
\textbf{VT-Former$_{LH}$}                 & 34.88                         & 100.59                        & 6.71        & 17.24       & 28.70       & 45.86       & 82.00          \\
\rowcolor[HTML]{D9D9D9} 
\textbf{VT-Former$_{MH}$}                 & 27.44        & 85.45        & 5.19        & 12.90        & 21.38       & 35.09       & 68.60          \\
\rowcolor[HTML]{D9D9D9} 
\textbf{VT-Former$_{SH}$}                 & 21.86        & 66.28        & 5.42        & 12.69       & 18.81       & 26.67       & 53.05          \\

\bottomrule
\end{tabular}%
}
\end{table}
\subsection{Comparison with State-of-the-Art Methods}
In this section, we conduct a comparative analysis between VT-Former and other trajectory prediction methods, leveraging three extensive highway datasets discussed in \cref{sec:datasets}: NGSIM \cite{NGSIM_US101, NGSIM_i80}, CHD High-angle \cite{chd} and CHD Eye-level \cite{chd}. Please note that VT-Former is only compared to deterministic models \cite{alinezhad2023pishgu, li2019grip++, katariya2022deeptrack} and not models such as \cite{messaoud2020attention} that predict multiple future trajectories. Both CHD datasets were recently released, and as such, the majority of SotA models have not been evaluated on this specific dataset. To bridge this gap, we have used the available code repositories of leading SotA models, reconfiguring them for training and evaluation on the CHD datasets.

\begin{figure}[]
    \centering
    % \resizebox{1\linewidth}{!}{
    \includegraphics[clip,trim={17 23 23 18},width=1\columnwidth]{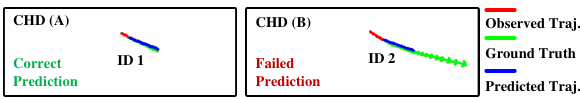}
    % }
    \caption{Qualitative performance of VT-Former on CHD, featuring a scene with a merging lane from the left and a mild right curve. The angle between predicted, observed, and actual trajectories in samples A and B demonstrates VT-Former's ability to predict the correct path with respect to road geometry. However, it struggles with the high acceleration of vehicle 2 in sample B. Samples are cropped for clarity.}
    \label{fig:qual}
\end{figure}

As discussed in \cref{sec:introduction}, this exploratory study assesses the advantages and limitations of using transformers and graphs for SVTP. The results presented in \cref{tab:ngsim3}, \cref{tab:chd_h3}, and \cref{tab:chd_e3}, along with the qualitative findings such as \cref{fig:qual}, collectively illustrate the potential of the VT-Former for highway trajectory prediction. Although the VT-Former did not achieve high results on the NGSIM dataset, it performed notably well as the second-best model on both the CHD High-angle and CHD Eye-level datasets. These findings indicate that while Graph Isomorphism Networks (GIN) and transformers are viable for SVTP, there is still potential for further refinement. More specifically, the data can lead to the hypothesis that the complexity of the graph significantly impacts the outcomes. Therefore, practices such as transitioning from fully connected graphs to more dynamic and optimized structures, along with implementing techniques such as neighborhood selection, could enhance performance.

\subsection{Observation Horizon Analysis}

\label{sec:OHA}

In this section, we conduct a comprehensive investigation into the influence of different observation horizons to discern the optimal configuration for each dataset. The observation horizon is varied, including 5 timesteps for VT-Former$_{SH}$ (translating to 1 second for 5Hz), 10 timesteps for VT-Former$_{MH}$ (translating to 2 second for 5Hz) 15 timesteps for VT-Former$_{LH}$ (translating to 3 second for 5Hz). Subsequently, we train and assess the performance of VT-Former on the NGSIM \cite{NGSIM_US101, NGSIM_i80}, CHD High-angle \cite{chd}, and CHD Eye-level \cite{chd} datasets. The results are presented in \cref{tab:ngsim3}, \cref{tab:chd_h3}, and \cref{tab:chd_e3} for the respective datasets. This specific assessment is especially important for applications that require faster prediction, such as collision avoidance or anomaly detection.

% \hl{In the context of the NGSIM dataset, we have observed variations in the accuracy of predicted trajectories when employing different observation horizons. These variations are at most $7.5\%$ and $6.1\%$ degradation in terms of ADE and FDE, respectively, as compared to the conventional 3-second observation setup. This particularly shows that even with a lower observation horizon, VT-Former is still able to accurately predict future trajectories. In the CHD High-angle dataset, a similar trend is observed, with the accuracy of predicted trajectories displaying consistency and minimal disparities among various observation horizons.}

The results on all three datasets, as depicted in \cref{tab:ngsim3}, \cref{tab:chd_h3}, and \cref{tab:chd_e3}, offer insights into the benefits of short-term observation horizon. Here, it is evident that a shorter observation horizon is more advantageous. For example, on the CHD Eye-level dataset, when deploying VT-Former$_{SH}$, we observe a remarkable enhancement of $37.3\%$ in FDE and $34.1\%$ in ADE compared to VT-Former$_{LH}$ results. This outcome underscores the practicality of employing shorter observation horizons in the eye-level setup. Similar trends are seen for both NGSIM and CHD High-angle datasets. This phenomenon can be attributed to the nature of trajectory prediction. In longer trajectories, maintaining a high temporal resolution (detail at very small time scales) might be challenging due to computational constraints or model architecture limits. This loss of detail can lead to poorer predictions as the model might miss out on subtle cues that are more apparent in shorter, high-resolution trajectories. This particularly shows that with a lower observation horizon, VT-Former is able to predict future trajectories better.

In summary, the empirical evidence presented in this study shows that, in practical, real-world scenarios, the adoption of shorter observation horizons is an effective strategy for decreasing the negative impact of chaotic driving behavior. This approach not only leads to more accurate trajectory prediction outcomes but also offers the advantage of expeditious prediction. By reducing the temporal extent of observation, the model can effectively focus on details at smaller scales, maintaining a high temporal resolution, thus enhancing the precision of its predictions and enabling it to respond more swiftly. This finding underscores the practical utility of optimizing observation horizons in trajectory prediction models, particularly in applications where the timely and precise forecasting of object movements is important.

%% file: tex/ablation.tex
\section{Ablation Study}
In this section, we perform ablation testing to evaluate and compare the effect of higher data rate on the performance of the model, including the Observation Hozrion test same as \cref{sec:OHA}. We utilize the NGSIM dataset as it offers a 10Hz sampling rate compared to conventional 5Hz sampling.
% As discussed, the output of VT-Former is compared with the actual trajectory to classify each trajectory as normal or anomalous. As shown in \cref{fig:anomaly}, the results demonstrate an AUC-ROC of 0.71 and AUC-PR of 0.82 with an EER of 0.32. The results of the driving anomaly detector demonstrate moderately effective performance, with an AUC-ROC score of 0.71, indicating a reasonable ability to distinguish anomalies from non-anomalies. In contrast, the AUC-PR score of 0.82 reflects a strong balance between precision and recall in identifying anomalies. The EER at 0.32 signifies a specific threshold point where the detector equally balances false positives and false negatives. These findings suggest that the model is proficient in identifying anomalies and minimizing false positives, making it well-suited for applications where precision is crucial, such as driving anomaly detection in security or maintenance contexts. Fine-tuning the operating threshold may further optimize the model's performance to align with specific application requirements.
% \subsection{Comparative Analysis for Shorter Prediction}
The performance of VT-Former and Pishgu \cite{alinezhad2023pishgu} with an observation horizon of 15 timesteps (translating to 1.5 seconds for 10Hz) and a prediction horizon of 25 timesteps (translating to 2.5 seconds for 10Hz) on the NGSIM dataset is presented in \cref{tab:ablation}. VT-Former shows a $7.95 \%$ improvement in ADE and $8.16 \%$ improvement in FDE over Pishgu. A similar trend is observed in RMSE with the most pronounced improvement of $7.47 \%$ from Pishgu's 1.74 meters to 1.61 meters over a 2.5-second prediction horizon. This shows that at a higher data rate, VT-Former not only outperforms Pishgu but achieves higher accuracy with respect to RMSE at further timesteps.

\begin{table}[]
\centering
\caption{Comparision of VT-Former and Pishgu on NGSIM dataset sampled at 10 Hz. VT-Former results are shown with different observation horizons of 15, 10, and 5 for VT-Former$_{LH}$, VT-Former$_{MH}$, and VT-Former$_{SH}$ respectively. Pishgu results are from its updated repository. The results are in meters.}
\label{tab:ablation}
\resizebox{\columnwidth}{!}{%
\begin{tabular}{@{}ccc|ccccc@{}}
\toprule
\multicolumn{1}{l}{}   & \multicolumn{1}{l}{} & \multicolumn{1}{l}{} & \multicolumn{5}{|c}{\textbf{RMSE}}                                   \\ \midrule
\textbf{Model}         & \textbf{ADE}         & \textbf{FDE}         & \textbf{0.5s} & \textbf{1.0s} & \textbf{1.5s} & \textbf{2.0s} & \textbf{2.5s} \\
\midrule
\textbf{Pisghu \cite{alinezhad2023pishgu}}                       & 0.88         & 1.96         & 0.15        & 0.46        & 0.82        & 1.25   & 1.74   \\
\midrule
\rowcolor[HTML]{D9D9D9}
\multicolumn{1}{c}{\textbf{VT-Former$_{LH}$}} & 0.81         & 1.80         & 0.14        & 0.42        & 0.76        & 1.16        & 1.61          \\
\rowcolor[HTML]{D9D9D9}
\multicolumn{1}{c}{\textbf{VT-Former$_{MH}$}} & 0.85         & 1.89         & 0.16        & 0.47        & 0.84        & 1.27        & 1.75      \\
\rowcolor[HTML]{D9D9D9}
\multicolumn{1}{c}{\textbf{VT-Former$_{SH}$}} & 0.79         & 1.78         & 0.13        & 0.41        & 0.76        & 1.16        & 1.62          \\
\bottomrule
\end{tabular}%
}
\end{table}

% \subsection{Varying Observation for Shorter Prediction}
Additionally, the impact of varying observation horizons, including 5 timesteps for VT-Former$_{SH}$ (translating to 0.5 seconds for 10Hz), 10 timesteps for VT-Former$_{MH}$ (translating to 1 second for 10Hz) 15 timesteps for VT-Former$_{LH}$ (translating to 1.5 seconds for 10Hz), is assessed and detailed in \cref{tab:ablation}. These results exhibit similar trends to those discussed in \cref{sec:OHA}, where a shorter observation horizon correlates with improved performance in predicting future trajectories.

%% file: tex/conclusion.tex
\section{Conclusion}

This work introduced VT-Former, an innovative trajectory prediction algorithm that leverages transformer architectures in conjunction with a novel tokenization scheme utilizing graph attention mechanisms denoted as Graph Attentive Tokenization (GAT). Extensive experimentation was conducted across three benchmark datasets illustrating the performance of the model, as well as assessing the model's generalizability and robustness. Further experimentation has been conducted to evaluate VT-Former's performance under shorter observed horizons. This study underscores the potential of combining graphs with transformers, opening a new path for future research in the field.